\documentclass{article}

     \PassOptionsToPackage{numbers, compress}{natbib}

     \usepackage[preprint]{submitted}

\usepackage[utf8]{inputenc} 
\usepackage[T1]{fontenc}    
\usepackage{hyperref}       
\usepackage{url}            
\usepackage{booktabs}       
\usepackage{amsfonts}       
\usepackage{nicefrac}       
\usepackage{microtype}      
\usepackage{xcolor}         

\usepackage{graphicx}
\usepackage{booktabs}
\usepackage{multirow}
\usepackage{arydshln}
\usepackage{subfigure}
\usepackage{amssymb} 
\usepackage{algorithm}
\usepackage{algpseudocode}
\usepackage{colortbl} 
\usepackage{arydshln} 
\usepackage{multicol}
\usepackage{amsthm}
\usepackage{arydshln}

\usepackage{amsthm}
\usepackage{makecell}
\usepackage{boldline}
\usepackage{color}
\usepackage{amsmath}
\usepackage{listings}
\usepackage{tcolorbox}
\usepackage{wrapfig}

\newtheorem{Thm}{Theorem}
\newtheorem{Def}{Definition}
\newtheorem{Cor}{Corollary}

\title{Top-$k$ Regularization for Supervised Feature Selection}

\author{%
  Xinxing Wu,
  Qiang Cheng\thanks{Corresponding author}\\
  University of Kentucky, Lexington, Kentucky, USA\\ 
  \texttt{qiang.cheng@uky.edu}\\
}

\begin{document}

\maketitle

\begin{abstract}
Feature selection identifies subsets of informative features and reduces dimensions in the original feature space, helping provide insights into data generation or a variety of domain problems. Existing methods mainly depend on feature scoring functions or sparse regularizations; nonetheless, they have limited ability to reconcile the representativeness and inter-correlations of features. In this paper, we introduce a novel, simple yet effective regularization approach, named top-$k$ regularization, to supervised feature selection in regression and classification tasks. Structurally, the top-$k$ regularization induces a sub-architecture on the architecture of a learning model to boost its ability to select the most informative features and model complex nonlinear relationships simultaneously. Theoretically, we derive and mathematically prove a uniform approximation error bound for using this approach to approximate high-dimensional sparse functions. Extensive experiments on a wide variety of benchmarking datasets show that the top-$k$ regularization is effective and stable for supervised feature selection.
\end{abstract}

\section{Introduction}\label{Intr}
Rapid advances in sensing technology have accelerated the use of high-dimensional data in diverse fields, ranging from social networks to high-throughput multi-omics in biomedicine~\citep{Cai,Katerine,LiData}. Generally, dimensionality reduction is necessary before in-depth analysis of such data. Feature selection, as a key dimension reduction technique, intends to identify the most informative features and discard irrelevant ones. A relevant yet different technique is feature extraction, such as principal component analysis (PCA)~\citep{Pearson} and auto-encoder (AE)~\citep{David,Ballard}; it projects the original features into a low-dimensional latent space, where the interpretation of the projected features is indirect. Feature selection, on the other hand, directly gleans insights of the original features~\citep{Guyon} and has wide applications, for instance, in the facilitation of deep learning model interpretation~\citep{Chen}, gene expression analysis~\citep{LiuGene}, and depression detection~\citep{Sharifa}.

Feature selection may be categorized into supervised, semi-supervised, and unsupervised methods according to whether or how label information is involved~\citep{Jennifer,Alelyani}. For supervised methods where labels are utilized, various regularization techniques, either deterministic or stochastic, have been studied. Least absolute shrinkage and selection operator (Lasso)~\citep{Tibshirani1996,Chen1998} is popular~\citep{Ma,Tang,Jovic,Li}, exploiting the sparsity of feature weights via the $\ell_1$-norm as a convex relaxation of the $\ell_0$-norm for feature selection. A large quantities of methods based on Lasso-type regularizations have been developed, e.g.,~\citep{Jun,Hara,Makoto}. In a spirit similar to the $\ell_1$-norm in Lasso-type methods,~\citet{Yamada} have recently introduced a probabilistic relaxation of the $\ell_0$-norm called Stochastic Gates (STGs) that relies on a continuous relaxation of Bernoulli distribution. Another probabilistic technique leverages the concrete random variables~\citep{Maddison} through deploying a concrete selector layer for unsupervised feature selection~\citep{Abubakar} or adding a supervision branch to the concrete selector layer for supervised feature selection~\citep{Singh}. 

Despite notable successes, these regularization methods mainly focus on selecting of features with large weights or probabilities; nonetheless, they pay little  attention to the reconciliation of the representativeness and correlations of features, thus leading to inconsistent or suboptimal performance with selected features. For example, Lasso regularization may select one from a group of correlated features and cause inconsistency in feature selection~\citep{Yuan}; also, Lasso-type regularizations typically lead to shrinking of feature weights~\citep{Zou2005}. Moreover, setting a threshold on the weights to select informative features may be problematic, as those features with weights (probabilities) below the threshold are ignored even if they may contribute to downstream learning models as well. This issue is clearly manifested in our experiments and will be demonstrated in Section~\ref{exp}. On the other hand, recent studies on extending Lasso-type regularizations to deep learning by focusing more on feature correlations, e.g., Deep Feature Selection (DFS)~\citep{Yifeng}, reported insufficient sparsity of the weights, resulting in loss of performance~\citep{Feng,Yamada}. Therefore, how to properly reconcile the representativeness and correlations of features is a challenge for feature selection as well as supervised learning. 

To address this challenge, we propose a novel regularization approach, named top-$k$ regularization, for supervised feature selection. It facilitates a sensible reconciliation of the representativeness and inter-feature correlations by selecting $k$ most informative features to optimize the predictive ability of existing learning models. Structurally, the proposed approach is concise through inducing a sub-architecture on an existing (lead) architecture, which is similar in construction yet has a potential to boost its predictive ability. Computationally, the top-$k$ regularized model is as efficient as the original lead model, where the learning of the sub-architecture can directly use the existing computation for the lead model with a slight modification. Theoretically, we obtain a uniform error bound for learning models with such a regularization to approximate high-dimensional sparse functions. In summary, our main contributions include the following:
\begin{itemize}
\item We propose an innovative regularization for supervised feature selection. It can be efficiently plugged as a sub-model
into a supervised learning model which serves as a lead model. It can ensure to sensibly reconcile the representativeness and correlations conditioning on $k$ selected features, potentially boosting the predictive ability of the lead models with the selected features.
\item In contrast to existing models that require nonlinear activation or link functions to enable nonlinearity, the proposed top-$k$ regularization naturally achieves nonlinear mapping even if the lead models are linear. 
\item We theoretically prove a uniform approximation error bound for models with this new regularization to approximate high-dimensional sparse functions, thereby providing a theoretical certificate of using this regularization for feature selection. 
\item We perform extensive experiments for regression and classification tasks, demonstrating the effectiveness of this regularization on $16$ benchmarking datasets, including $1$ simulation and $15$ real datasets. The top-$k$ regularization also demonstrates more stable performance on varying numbers of selected features than strong baseline methods.
\end{itemize}

Let $n$, $m$, and $k$ be the numbers of samples, features, and selected features, respectively. $\mathbf{X}\in\mathbb{R}^{n\times m}$ is the feature matrix of $n$ examples, and $\mathbf{y}\in\mathbb{R}^n$ contains their responses or labels. Besides, $\mathbf{w}\in\mathbb{R}^m$ is a vector of feature weights, $\mathbf{w}^{\mathrm{max}_k}$ is defined to be an operator to keep the $k (\leqslant m)$ largest entries of $\mathbf{w}$ in magnitude while making the other entries $0$; $\mathbf{w}^{k}$ denotes an operator to obtain a sub-vector by picking up $k (\leqslant m)$ entries from $\mathbf{w}$; $\mathrm{Diag}(\mathbf{w})$ is a diagonal matrix with the diagonal vector $\mathbf{w}$.

\section{Related work}
Here we will briefly review the formalization of feature selection and discuss typical regularization techniques for feature selection.

Feature selection methods are often categorized into four approaches, including filter, wrapper, embedder, and hybrid approaches~\citep{Alelyani,Li}.  We will focus on the embedder approach that embeds feature selection into other learning models to achieve model fitting and feature selection simultaneously.

\paragraph{Formalization of feature selection.}

The problem of selecting a subset of features with supervision can be stated as follows: for a fixed positive integer $k$, 
\begin{equation}\label{fs}
\max_{S^k}Q(\mathbf{X}_{S^k},\mathbf{y}),
\end{equation}
where $S^k$ denotes a subset of $k$ features, $\mathbf{X}_{S^k}$ is the derived data set from $\mathbf{X}$ based on $S^k$, and $Q$ denotes the feature selection quality function, whose optimization will take into account $\mathbf{y}$ and $\mathbf{X}_{S^k}$. The larger the value of $Q$, the better. With an embedder approach, the objective function of the learning model to embed with feature selection can serve as a proxy for $Q$. For example, for regression, the negative mean absolute error (MAE) or negative mean least squares (MSE) may serve as $Q$; for classification, the accuracy or negative hinge loss  may serve as $Q$.

\paragraph{$\ell_0$-norm.}

In the linear setting with a least squares loss,~\eqref{fs} reduces to:
\begin{equation}{\label{l0}}
\min_{\mathbf{w}\in\mathbb{R}^{m}}\|\mathbf{y}-\mathbf{X}\mathbf{w}\|^2_2\,\,s.t.\,\,\|\mathbf{w}\|_0=k,
\end{equation}
where $\|\mathbf{w}\|_0$ is a pseudo-norm measuring the number of nonzero elements of $\mathbf{w}$. The exact optimization of~\eqref{l0} needs to search the space of all possible subsets, which is a NP-hard problem~\citep{Natarajan,Weston,Hamo}. A variety of surrogate functions for $\ell_0$-norm have been developed, including regularization techniques mentioned below. The proposed regularization is a direct counterpart of the $\ell_0$ norm; notably, it can be easily plugged in and efficiently optimized along with lead models. 

\paragraph{$\ell_1,\ell_2,\ell_1+\ell_2$-norms.}

Lasso~\citep{Tibshirani1996,Chen1998} leverages the $\ell_1$ norm for regularization and feature selection to enhance the prediction accuracy and interpretability of the linear regression model. It selects only one from a group of correlated features~\citep{Yuan}. Despite good representativeness, the $\ell_1$-norm takes no account of group structures or inter-correlation of features, which potentially have adverse effect on the predictive ability. To rectify this limitation, elastic net~\citep{Zou2005} uses the $\ell_1+\ell_2$ norm to combine the grouping effect of the $\ell_2$-norm~\citep{Hoerl1970} with the sparsity of Lasso:
\begin{equation}{\label{Elastic}}
\min_{\mathbf{w}\in\mathbb{R}^{m}}\|\mathbf{y}-\mathbf{X}\mathbf{w}\|^2_2+\lambda_2\|\mathbf{w}\|_1+\lambda_1\|\mathbf{w}\|_2,
\end{equation}
where~\eqref{Elastic} reduces to Lasso if $\lambda_1=0$, and to ridge regression if $\lambda_2=0$. With nonzero $\lambda_1$ and 
$\lambda_0$, it encourages sparse effect at the group level, with correlated features selected as a group.

Elastic net-type models tend to select a whole group of strongly correlated features, thus leading to potential redundancy in many tasks. They have good inter-feature correlations yet may lack good representativeness for promoting predictive ability. In contrast, Lasso-type regularizations yield good representativeness without inter-correlations as previously discussed. A sensible reconciliation of the representativeness and inter-correlations of features is yet to be achieved. Considering these limitations, we argue that the learning ability of the lead models with the selected features will constitute for a more sensible quality measure in~\eqref{fs} to determine whether correlated features should be selected compared to the sparsity and/or group effect. In this paper, we will exploit this idea to introduce a regularization approach to achieve a necessary and well-controlled reconciliation of the representativeness and inter-feature correlations, thereby boosting downstream learning models. 

{\bf{Stochastic counterparts}} of the above regularization techniques~\citep{Abubakar,Yamada} are available as mentioned in Section~\ref{Intr}. In experiments, we will compare the proposed regularization with two current state-of-the-art methods with probabilistic constraints/regularizations, Concrete Selector Feature Selection (CS-FS)~\citep{Abubakar}{\footnote{For regression analysis, we replace the original output $\mathbf{x}$ with $y$ for supervised feature selection.}} and Stochastic Gates Feature Selection (STG-FS)\citep{Yamada}. 

Additionally, for thorough assessment of our approach, we will also compare with other feature selection methods, including a deep extension of~\eqref{Elastic}, DFS, an often used ensemble method, Random Forest (RF)~\citep{Breiman}, and two contemporary methods, graph-based Infinite Feature Selection (Inf-FS)~\citep{Giorgio} and Feature Importance Ranking for Deep Learning (FIR-DL)~\citep{Wojtas}. 

\section{Method}\label{meth}
Here we will present the top-$k$ regularization, its use cases in supervised feature selection and theoretical analysis of its approximation ability.

\paragraph{Top-$k$ regularization.}

We propose this innovative approach to reconcile the representativeness and correlations of features in supervised feature selection. It is a direct counterpart of the $\ell_0$-norm regularization. We leverage the predictive ability of a given learning model, called a lead model, with a subset of selected $k$ features to form a regularization term. Structurally, this approach uses the operator $\mathbf{w}^{\mathrm{max}_k}$ to pinpoint $k$ most informative features and introduce a sub-architecture into the main architecture of the lead model. The induced sub-architecture has a similar fitting error term to the lead model, and is trained cooperatively with the main architecture. By using this regularization term, we speculate that the resulting model can select representative features and capture complex nonlinear inter-feature relationship simultaneously. As will be empirically and theoretically shown, this simple approach for regularization is able to select subsets of $k$ features to enhance the learning ability of lead models, achieving classification or regression performance on par with or better than strong baseline methods. 

We will show the use cases of the proposed regularization adopting Lasso, ridge, elastic net, and deep neural networks (DNNs) as lead models.

\paragraph{Top-$k$ regularization for Lasso.}

When Lasso is used as a lead model, we plug in a sub-model which is similar to Lasso structurally but takes only top-$k$ features as input. Concretely, we use the operator $\mathbf{w}^{\mathrm{max}_k}$ to form the top-$k$ regularization as follows: 
\begin{equation}{\label{lassoLagrangian_k}}
\min_{\mathbf{w}\in\mathbb{R}^{m}}\|\mathbf{y}-\mathbf{X}\mathbf{w}\|^2_2+\lambda_2\|\mathbf{y}-\mathbf{X}\mathbf{w}^{\mathrm{max}_k}\|^2_2+\lambda_1\|\mathbf{w}\|_1,
\end{equation}
where $\lambda_i$ are nonnegative hyper-parameters. The second term is for the top-$k$ regularization, which focuses on the predictive ability of selected features, i.e., simply requires the features with the highest weights in magnitude to fit the responses well. The optimization of~\eqref{lassoLagrangian_k} is similar to ordinary Lasso, which is typically iterative though there exist a number of ways for doing it. In each iteration, only a slight change is needed due to the introduction of the second term; that is, for the top-$k$ features at this iteration, the computation from the first term of~\eqref{lassoLagrangian_k}, e.g., gradients, can be re-used for the second term, while for other features there is no change compared to Lasso optimization. We consider a set of simulation data to illustrate the behaviors of~\eqref{lassoLagrangian_k} in Figure~1 (c) in the Supplementary Material. It is noted that Lasso identities $16$ informative features, whose weights, however, are shrunk of the true values; on the other hand,~\eqref{lassoLagrangian_k} identifies $20$ informative features, whose weights are more consistent than Lasso.

\paragraph{Top-$k$ regularization for elastic net and ridge regression.}

When elastic net (\ref{Elastic}) is used as a lead model, in a similar way to the Lasso case, we plug in a sub-model using $\mathbf{w}^{\mathrm{max}_k}$ to induce the top-$k$ regularization as follows:
\begin{equation}{\label{Elastic_top}}
\min_{\mathbf{w}\in\mathbb{R}^{m}}\|\mathbf{y}-\mathbf{X}\mathbf{w}\|^2_2+\lambda_3\|\mathbf{y}-\mathbf{X}\mathbf{w}^{\mathrm{max}_k}\|^2_2+\lambda_2\|\mathbf{w}\|_1+\lambda_1\|\mathbf{w}\|_2.
\end{equation}

Here, the second term is for the top-$k$ regularization. The parameters $\lambda_i$ control the strengths of different regularization terms. When $\lambda_1=0$,~\eqref{Elastic_top} reduces to~\eqref{lassoLagrangian_k}; when $\lambda_2=0$,~\eqref{Elastic_top} represents the ridge regression model with the top-$k$ regularization; when $\lambda_3 = 0$,~\eqref{Elastic_top} reduces to ordinary elastic net. The optimization of~\eqref{Elastic_top} is efficient, which is similar to the optimization of~\eqref{lassoLagrangian_k}. We run the analysis of~\eqref{Elastic_top} on the simulation data and illustrate the results in Figure~1 (e)-(h) in the Supplementary Material. It is observed that the top-$k$ regularization helps identify more informative features and yield more consistent weights than ordinary elastic net or ridge regression.

We also experiment elastic net, ridge regression, and their top-$k$ regularized versions on the testing set of the simulation data, and the results in Figure~1 (i)-(l) in the Supplementary Material indicate that the proposed regularization can effectively enhance these downstream learning models.

It is worth noting that, in contrast to existing models, including NNs and generalized linear models, which need nonlinear activation or link functions to model nonlinear relationships, the top-$k$ regularization naturally facilitates a nonlinear model even if the lead model is linear, such as~\eqref{Elastic}. The intrinsic nonlinearity of our regularization and models, such as~\eqref{Elastic_top}, stems from the weight ranking and selecting the $k$ top features during the iterative optimization.

\paragraph{Top-$k$ regularization for DNNs.}

Now we use DNNs as lead learning models and introduce the top-$k$ regularization into DNNs. For this purpose, we first revisit the first and second terms in~\eqref{lassoLagrangian_k} or~\eqref{Elastic_top} and treat them from a NN perspective. Indeed, we can regard them as special NNs with a one-to-one layer and a constant layer. Mathematically, they can be represented as $\min_{\mathbf{w}\in\mathbb{R}^{m}}\|\mathbf{y}-\mathbf{X}\mathrm{Diag}(\mathbf{w})\mathbf{1}\|^2_2+\lambda_3\|\mathbf{y}-\mathbf{X}\mathrm{Diag}(\mathbf{w}^{\mathrm{max}_k})\mathbf{1}\|^2_2$, where $\mathbf{1}$ is an $m\times 1$ vector whose elements are all $1$. From this perspective, we can plug in the top-$k$ regularization term by modifying the constant layer and extend~\eqref{Elastic_top} to DNNs by retaining the one-to-one layer as follows:
\begin{equation}{\label{DNNReg}}
\displaystyle\min_{\mathbf{w}\in\mathbb{R}^{m},\mathbf{F}}\|\mathbf{y}-\mathbf{F}(\mathbf{X})\|^2_2+\lambda_3\|\mathbf{y}-\mathbf{F}^{\mathrm{max}_k}(\mathbf{X})\|^2_2+\lambda_2\|\mathbf{w}\|_1+\lambda_1\|\mathbf{w}\|_2.
\end{equation}
Here, $\mathbf{F}$ (resp., $\mathbf{F}^{\mathrm{max}_k}$) denotes an $L$-layer (resp., sub-) NN with a one-to-one layer; mathematically, $\mathbf{F}(\mathbf{X})\triangleq\sigma_L\circ f_L\circ\cdots\sigma_1\circ f_1\circ(\mathbf{X}\mathrm{Diag}(\mathbf{w}))$ and $\mathbf{F}^{\mathrm{max}_k}(\mathbf{X})\triangleq\sigma_L\circ f_L\circ\cdots\sigma_1\circ f_1\circ(\mathbf{X}\mathrm{Diag}(\mathbf{w}^{\mathrm{max}_k}))$, where $f_i(\mathbf{z})=\mathbf{z}\mathbf{W}^{(i)},${\footnote{Without loss of generality, we include the biases in the weights.}} $i=1, \ldots, L-1$, $f_L(\mathbf{z})=\mathbf{z}\mathbf{w}^{(L)}$, $\mathbf{W}^{(i)}\in\mathbb{R}^{M_{i-1}\times M_i}$, $M_i$ denotes the output dimension of the $i$-th layer, and $\mathbf{w}^{(L)}\in\mathbb{R}^{M_{L-1}\times 1}$. When $\mathbf{F} (\mathbf{X}) = \mathbf{X}\mathrm{Diag}(\mathbf{w})){\bf{1}}$,~\eqref{DNNReg} simply reduces to~\eqref{Elastic_top}. From our experimental results, we find that adopting  the linear activation for the last layer of~\eqref{DNNReg} and the ReLU for other layers of $\mathbf{F}$ generally works well. 

The optimization of~\eqref{DNNReg} is an efficient extension of the lead DNN models, where the main difference comes from the second term of~\eqref{DNNReg}. In each iteration of back propagation, the second term of~\eqref{DNNReg} would require the gradients of the features having the top-$k$ weights in magnitude during that iteration, while having no effect on the gradients of other features. Note that the sub-model in the second term essentially has the same architecture as the lead model in the first term. Thus, the gradients of the second term can reuse (up to a multiplicative factor) those corresponding to the top-$k$ weights from the first term, with only an additional ranking operation. Finding the top $k$ out of $m$ weights in each iteration has a worst-case efficiency 
$\mathcal{O}(m\min\{\log m, k\})$ that is independent of $n$. Therefore, the efficiency of optimizing~\eqref{DNNReg} is essentially the same as the lead DNN model.

\paragraph{Theoretical analysis for top-$k$ regularization.}

To obtain insights into the approximation ability of the proposed regularization, we derive a theoretical analysis of the predictive error using $\mathbf{F}^{\max_k}$.
\begin{Def}[$k$-sparse function]
$\forall~\mathbf{H}:\mathbb{R}^m\rightarrow\mathbb{R}$, a set of $k$-sparse functions is defined as
\begin{equation}{\label{ksparse}}
\begin{array}{l}
\mathcal{H}_k=\{\mathbf{H} \,|\, \exists\overline{\mathbf{H}}:\mathbb{R}^k\rightarrow\mathbb{R},\,\mathrm{s.\,t.}~\overline{\mathbf{H}}(\mathbf{x}^k)=\mathbf{H}(\mathbf{x}), |\overline{\mathbf{H}}(\mathbf{x}^k)|\leqslant\eta,\,\mathrm{and} \, | \overline{\mathbf{H}}(\mathbf{x}^k)-\overline{\mathbf{H}}({\mathbf{x}'}^{k})|\\
\leqslant\frac{\eta\|\mathbf{x}^k-{\mathbf{x}'}^{k}\|_2}{R},\|\mathbf{x}^k\|\leqslant R,\|{\mathbf{x}'}^{k}\|\leqslant R,\mathbf{x}^k, {\mathbf{x}'}^{k}\in\mathbb{R}^k,\mathbf{x}\in\mathbb{R}^m,\eta\, \mathrm{and}\, R\, \mathrm{are\, constants}\}.
\end{array}
\end{equation}
\end{Def}
We also define a set of $L$-layer sub-NNs induced by the top-$k$ regularization below.
\begin{Def}[A set of top-$k$ regularization induced sub-NNs]
\[\displaystyle\mathcal{F}_{L}^{\mathrm{top}-k}=\displaystyle\{\mathbf{F}^{\mathrm{max}_k} \,|\, \mathbf{F}^{\mathrm{max}_k}(\mathbf{x})=\sigma_L\circ f_L\circ\cdots\sigma_1\circ f_1\circ(\mathbf{x}\mathrm{Diag}(\mathbf{w}^{\mathrm{max}_k})),\,\forall\mathbf{x}\in\mathbf{X}\}.
\]
\end{Def}

Next, we will theoretically analyze the approximation error by using $L$-layer sub-NNs in~$\mathcal{F}_{L}^{\mathrm{top}-k}$ to approximate $k$-sparse high-dimensional functions in $\mathcal{H}_k$.
\begin{Thm}[Top-$k$ error bound]{\label{Approximationerror}}
There exists a $\delta$ that is greater than a constant depending only on $k$, $\forall\mathbf{H}\in\mathcal{H}_k$, $\exists\mathbf{F}^{\mathrm{max}_k}\in\mathcal{F}_{2}^{\mathrm{top}-k}$, we have
\begin{equation}{\label{errorbound_1}}
\displaystyle\sup_{\mathbf{x}\in\mathbf{X}}\left|\mathbf{H}(\mathbf{x})-\mathbf{F}^{\mathrm{max}_k}(\mathbf{x})\right|\leqslant C(k)\eta\left({\delta}/{\eta}\right)^{-{2}/{(k+1)}} \log({\delta}/{\eta})+C(k)\delta Q^{-{(k+3)}/{2k}},
\end{equation}
where $Q=(M-2)/2$, $M$ is the total number of neurons in the hidden layers, $\eta$ is a (Lipschitz) constant for $\mathcal{H}_k$, and $C(k)$ is a constant depending on $k$.
\end{Thm}

\begin{Cor}\label{cor1}
Optimizing~\eqref{errorbound_1} over $\delta$, we have the following uniform error bound: $\sup_{\mathbf{x}\in\mathbf{X}}\left|\mathbf{H}(\mathbf{x})-\mathbf{F}^{\mathrm{max}_k}(\mathbf{x})\right|\leqslant{2C(k)\eta\log M'}/{\left(M'\right)^{1/k}}$.
\end{Cor}

Theorem~\ref{Approximationerror} establishes an approximation error bound for using the sub-NNs induced by the top-$k$ regularization for feature selection. The bound in~\eqref{errorbound_1} depends on the number of hidden neurons $M$ of NN and that of informative features $k$ being selected. With greater $M$, the approximation tends to be tighter by Corollary~\ref{cor1}. When the Lipschitz constant $\eta$ increases, the approximation tends to be loose. The proofs of Theorem~\ref{Approximationerror} and Corollary~\ref{cor1} are given in the Supplementary Material.

\paragraph{Top-$k$ regularization for classification.}

By replacing the MSE loss of~\eqref{DNNReg} with the cross-entropy loss, we obtain the corresponding top-$k$ regularized model for classification,
\begin{equation}{\label{DNNClassification}}
\begin{array}{l}
\displaystyle\min_{\mathbf{w}\in\mathbb{R}^{m},\mathbf{F}}\mathrm{CE}(\mathbf{y},\mathbf{F}(\mathbf{X}))+\lambda_3\mathrm{CE}(\mathbf{y},\mathbf{F}^{\mathrm{max}_k}(\mathbf{X}))+\lambda_2\|\mathbf{w}\|_1+\lambda_1\|\mathbf{w}\|_2,
\end{array}
\end{equation}
where $\mathrm{CE}$ denotes the binary or categorical cross-entropy loss. Other loss functions for classification may also be employed. In our experiments, we adopt the Sigmoid/Softmax activation for the last layer of~\eqref{DNNClassification} and ReLU for the  other layers of $\mathbf{F}$. The optimization of~\eqref{DNNClassification} is similar to~\eqref{DNNReg}. When the top-$k$ regularization is not in use, i.e., $\lambda_3=0$,~\eqref{DNNReg} and~\eqref{DNNClassification} reduce to DFS.

\section{Experiments}\label{exp}
We will perform extensive experiments to evaluate the proposed top-$k$ regularization and the resulting learning models by comparing them to strong contemporary methods. 

\paragraph{Datasets to be used.}

The benchmarking datasets used in this paper include APPA-REAL~\citep{Agustsson}, PARSE~\citep{Antol}, DrivFace~\citep{Katerine}, Pyrimidines~\citep{King}, Triazines~\citep{KingR}, USPS~\citep{Hull}, MNIST~\citep{LeCun}, MNIST-Fashion~\citep{MNISTFashion}, ALLAML~\citep{Li}, COIL20~\citep{Sammeer}, Yale~\citep{Cai}, Kinematics, Ailerons, CompAct, and CompActSmall.{\footnote{The last four datasets are downloaded from https://www.dcc.fc.up.pt/\textasciitilde ltorgo/Regression/DataSets.html}} We summarize their statistics in Table~\ref{datasets} and give the details of their preprocessing in the Supplementary Material.
\begin{table}[!htp]
\caption{Statistics of datasets (Tr: training; Te: testing).}
\label{datasets}
\begin{center}
\begin{small}
\setlength{\tabcolsep}{3mm}{
\begin{tabular}{l|ccc}
\hline\hline
     Dataset			&\# Sample				&\# Feature 						& Range /\# Class\\
    \hline
    Pyrimidines											& 74   					& 27       							&  $[0.1,0.9]$\\
    Triazines 											& 186    			    & 60       							&  $[0.1,0.9]$\\
    Kinematics											& 8,192   				& 8       							&  $[0.04,1.46]$\\
    Ailerons							    				& Tr: 7,154, Te: 6,596  & 40       							&  $[-0.0036,0.0]$\\
    CompAct							    				& 8,192   				& 21       							&  $[0.0,99.0]$\\
    CompActSmall											& 8,192    				& 12       							&  $[0.0,99.0]$\\
    APPA-REAL												& Tr: 5,613, Te: 1,978  & Inhomogeneous  					& $[1,100]$\\
    PARSE													& 305 					& Inhomogeneous 					& $[0.14,28.16]^{28}$\\
    DrivFace												& 606 					& 6,400  							& $[-45,45]$\\
    USPS			                            			& 9,298 			    & 256 				                & 10\\
    MNIST												& Tr: 60,000, Te: 10,000& 784								& 10\\
    MNIST-Fashion										& Tr: 60,000, Te: 10,000& 784								& 10\\
	Yale													& 165 					& 1,024 							& 15\\
	COIL20												& 1,440					& 1,024								& 20\\
	ALLAML												& 72					& 7,129								& 2\\
\hline\hline
\end{tabular}}
\end{small}
\end{center}
\end{table}

\paragraph{Evaluation metrics.}

Three metrics are adopted for evaluating feature selection methods: 1) F1-score, which is used the same way as~\citep{Yamada}: the precision and recall are respectively the fractions of the selected informative features with respect to the total numbers of selected and informative features; 2) MAE, which is measured by passing selected features to a downstream regression model to evaluate their predictive ability; 3) Accuracy, which is measured by passing the selected features to a downstream classifier to benchmark their discriminative ability. For fair comparison, in the regression setting, we calculate F1-score and use an ordinary linear regression model as a downstream model to obtain the MAE measure; in the classification setting, we use extremely randomized trees~\citep{Geurts} as a downstream classifier to obtain the Accuracy measure. 

All experiments are implemented with Keras 2.3.1, Tensorflow 1.15.0, Scikit-Learn 0.24.2, Python 3.7.8, and JupyterLab 2.2.9. The parameter/experiment settings and the codes can be found in the Supplementary Material. All codes will be made publicly available upon acceptance. 

\paragraph{Results for regression.}

We randomly choose $20$ samples and calculate their mean and standard deviation; then we generate an independent Gaussian distribution using $0.1$ times the calculated mean and $0.01$ times the standard deviation. We sample this Gaussian distribution to produce as many noise features as the original features, then we concatenate the noise with the original features. The resulting dataset has the same sample size as the original dataset while twice the number of original features. After injecting noise into the 6 datasets in this way, we apply different feature selection methods on the noisy datasets. The original features are regarded to be truly informative for these relatively low-dimensional datasets to calculate the F1-scores. The results in Table~\ref{resultsReg1} indicate that the regression models with the top-$k$ regularization exhibit performance on par with or better than the strong baselines in detecting informative features.{\footnote{Here, $k$ is set to the number of original features for each dataset.}} With or without the top-$k$ regularizations, elastic net exhibits significantly better results than DNN on the two datasets of Triazines and Ailerons. A possible reason is that the input-output relationships of these two datasets are relatively simple and thus elastic net would suffice to model them. Nonetheless, on datasets with more complex relationships, DNN with the proposed regularization generally outperforms baseline methods. We also provide the detection results for visualization in the Supplementary Material.
\begin{table}[!thp]
\caption{Comparison of feature selection methods in F1-score for regression tasks.}
\label{resultsReg1}
\begin{center}
\begin{small}
\setlength{\tabcolsep}{2mm}{
\begin{tabular}{lcccccccccccc}
\hline\hline
Dataset	        & Pyrimidines & Kinematics &CompAct& Triazines & Ailerons &CompActSmall       \\
\hline
	Elastic Net					 	 &51.9		   &62.5		   &47.6	&46.7		   &62.5		 &50.0		 \\
    DFS  		    			 	 &81.5		   &{\bf 75.0}     &47.6	&43.3		   &27.5		 &41.7		 \\
    RF								 &7.4		   &62.5		   &9.5	    &3.3	   	   &10.0		 &8.3		 \\
	CS-FS  		    			 	 &74.1		   &25.0		   &47.6	&48.3		   &42.5		 &8.3		 \\
	Elastic Net$^{\mathrm{top}}$ 	 &51.9		   &62.5		   &47.6	&{\bf 78.3}	   &{\bf 82.5}	 &50.0		 \\
	DNN$^{\mathrm{top}}$ 		     &{\bf 88.9}  &{\bf 75.0}      &{\bf 81.0} &50.0       &45.0		  &{\bf 58.3}\\	
\hline\hline
\end{tabular}}
\end{small}
\end{center}
\end{table}

\begin{figure}[!thp]
\begin{center}
{
\begin{minipage}[t]{0.2\linewidth}
\centerline{\includegraphics[width=2.4\textwidth]{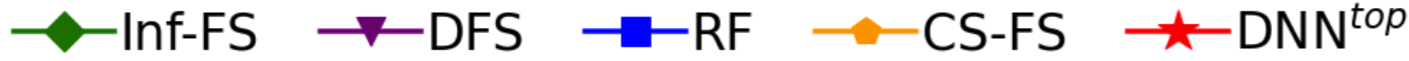}}
\end{minipage}%
}%
\hspace{5in}
\subfigure[DrivFace]{
\centering
{
\begin{minipage}[t]{0.3\linewidth}
\centering
\centerline{\includegraphics[width=0.9\textwidth]{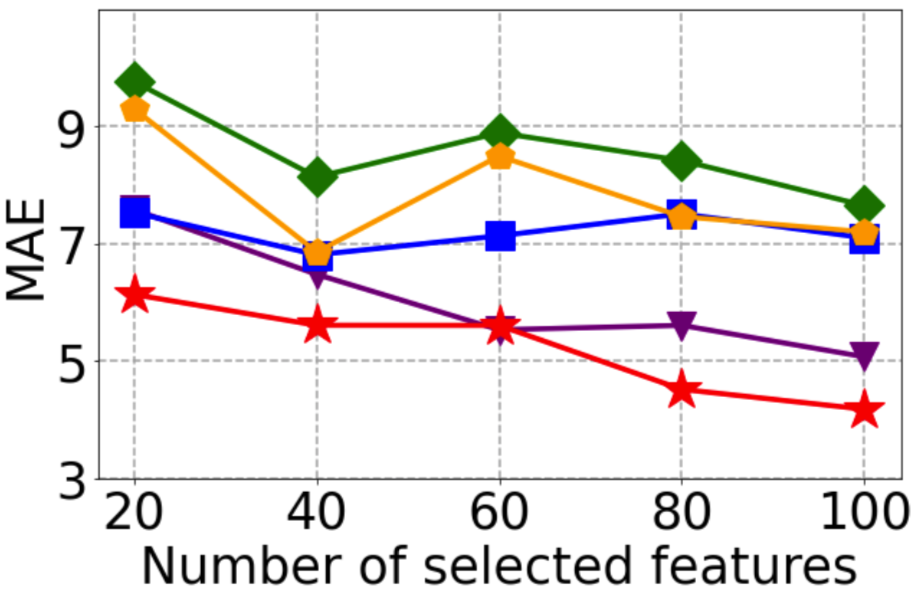}}
\end{minipage}%
}%
}%
\hspace{0.1in}
\subfigure[PARSE]{
\centering
{
\begin{minipage}[t]{0.3\linewidth}
\centering
\centerline{\includegraphics[width=0.95\textwidth]{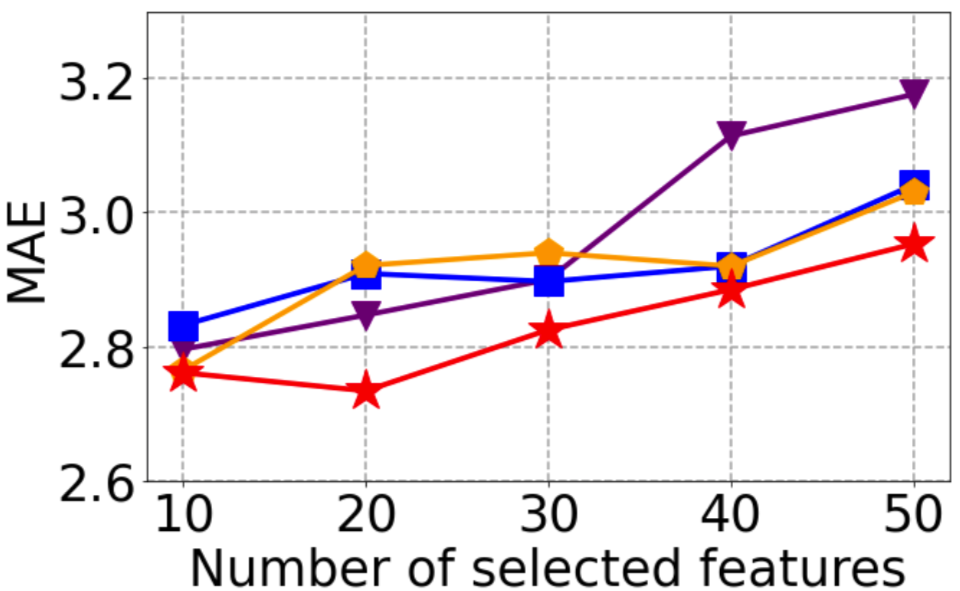}}
\end{minipage}%
}%
}%
\hspace{0.1in}
\subfigure[APPA-REAL]{
\centering
{
\begin{minipage}[t]{0.3\linewidth}
\centering
\centerline{\includegraphics[width=0.92\textwidth]{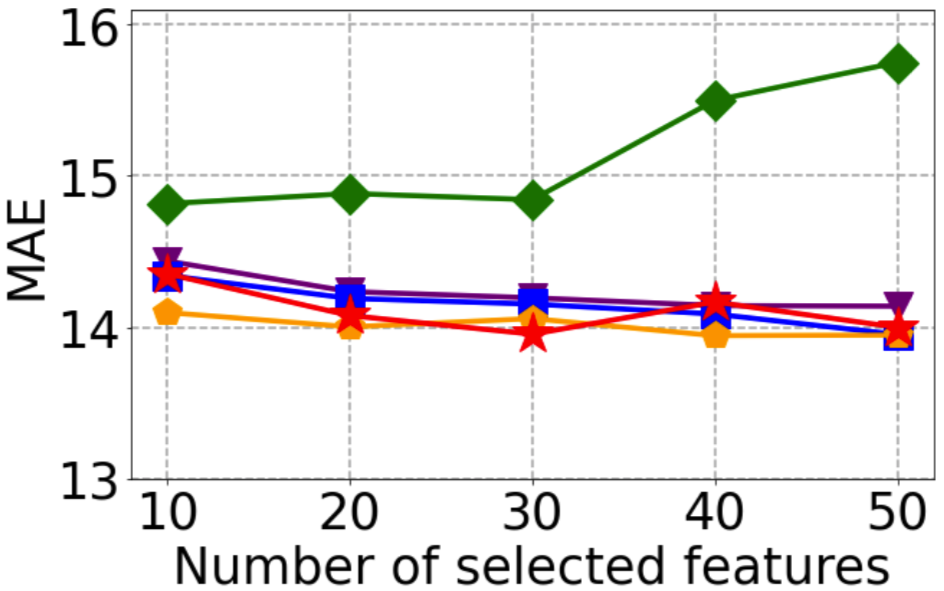}}
\end{minipage}%
}%

}%
\end{center}
\caption{MAE versus $k$ for regression. Smaller is better.}
\label{regression2_results}
\end{figure}

Further, we compare the behaviors of the DNN model with the top-$k$ regularization versus $k$ to those of the baselines. We vary $k$ from $10$ to $50$ with a step size of $10$ for PARSE and APPA-REAL and from $20$ to $100$ with a step size of $20$ for DrivFace. The plots of MAEs versus the number of selected features are presented in Figure~\ref{regression2_results}. The averaged results for different $k$'s are summarized in Table~\ref{resultsReg2},{\footnote{In these and subsequent experiments, as the remaining datasets have higher dimensions and potential nonlinearity, we will not use the elastic net-based model; instead, we use~\eqref{DNNReg} and~\eqref{DNNClassification} with $L=3$.}} indicating that~\eqref{DNNReg} outperforms baseline methods in most cases. 

\begin{table}[!thp]
\vspace{-0.25em}
\caption{Comparison of feature selection methods in averaged MAE.}
\label{resultsReg2}
\begin{center}
\begin{small}
\setlength{\tabcolsep}{6mm}{
\begin{tabular}{lccc}
\hline\hline
Dataset							&DrivFace     						&PARSE	   		&APPA-REAL   \\
\hline
	DFS							&6.05$\pm$0.15					&2.97$\pm$0.15 &14.23$\pm$0.11\\
    RF  		                &7.06$\pm$0.35 			  		&2.92$\pm$0.07     	&14.14$\pm$0.13       \\
	CS-FS                       &7.86$\pm$0.90 			  		&2.92$\pm$0.09     	&{\bf 14.01$\pm$0.06}     	\\
	Inf-FS						&8.57$\pm$0.72  			  	&    -			  	&15.16$\pm$0.39   	\\
	DNN$^{\mathrm{top}}$ 		&{\bf 5.21$\pm$0.74}   			&{\bf 2.83$\pm$0.08}&14.11$\pm$0.14     	\\
\hline\hline
\end{tabular}}
\end{small}
\end{center}
\vspace{-0.8em}
\end{table}

\paragraph{Results for classification.}

\begin{figure}[!thp]
\begin{center}
{
\begin{minipage}[t]{0.2\linewidth}
\centerline{\includegraphics[width=2.4\textwidth]{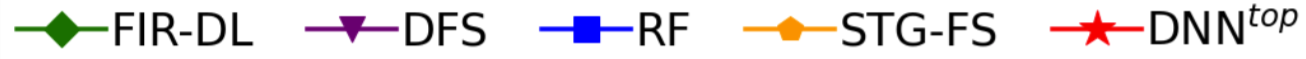}}
\end{minipage}%
}%
\hspace{5in}
\subfigure[Yale]{
\centering
{
\begin{minipage}[t]{0.3\linewidth}
\centering
\centerline{\includegraphics[width=0.9\textwidth]{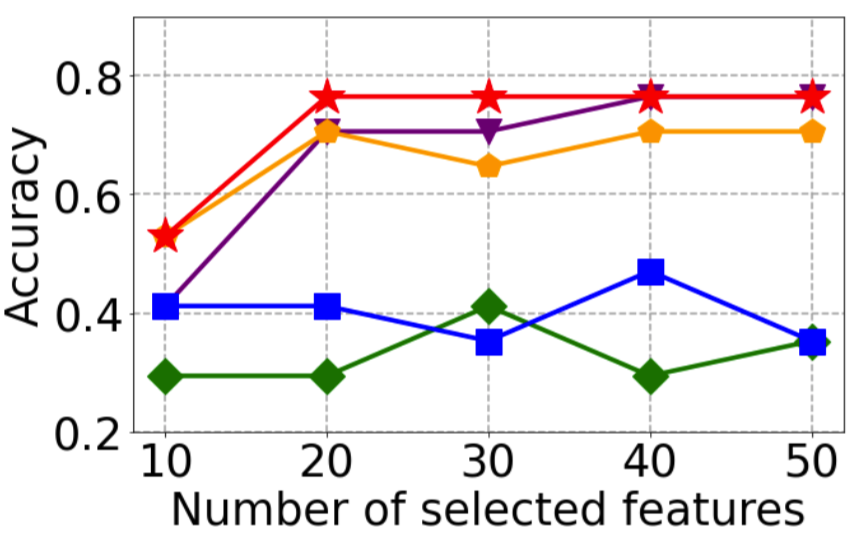}}
\end{minipage}%
}%
}%
\hspace{0.06in}
\subfigure[MNIST]{
\centering
{
\begin{minipage}[t]{0.3\linewidth}
\centering
\centerline{\includegraphics[width=0.9\textwidth]{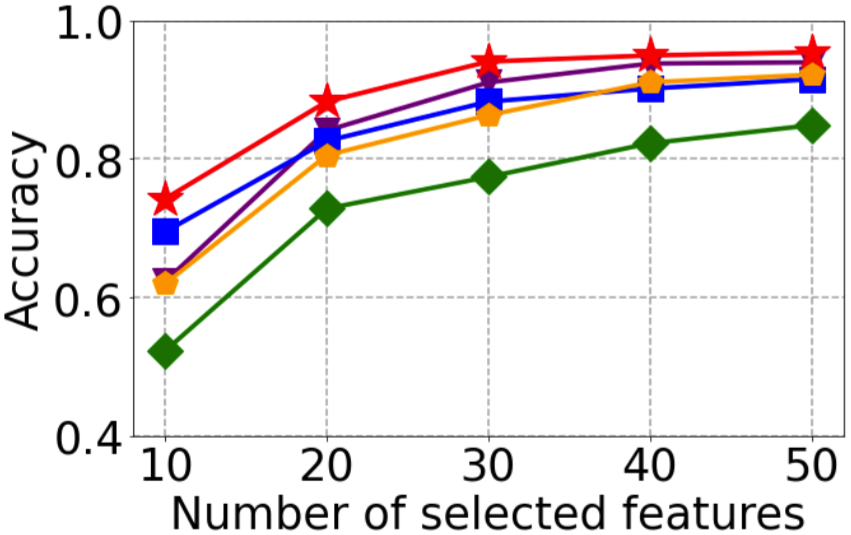}}
\end{minipage}%
}%
}%
\hspace{0.06in}
\subfigure[MNIST-Fashion]{
\centering
{
\begin{minipage}[t]{0.3\linewidth}
\centering
\centerline{\includegraphics[width=0.9\textwidth]{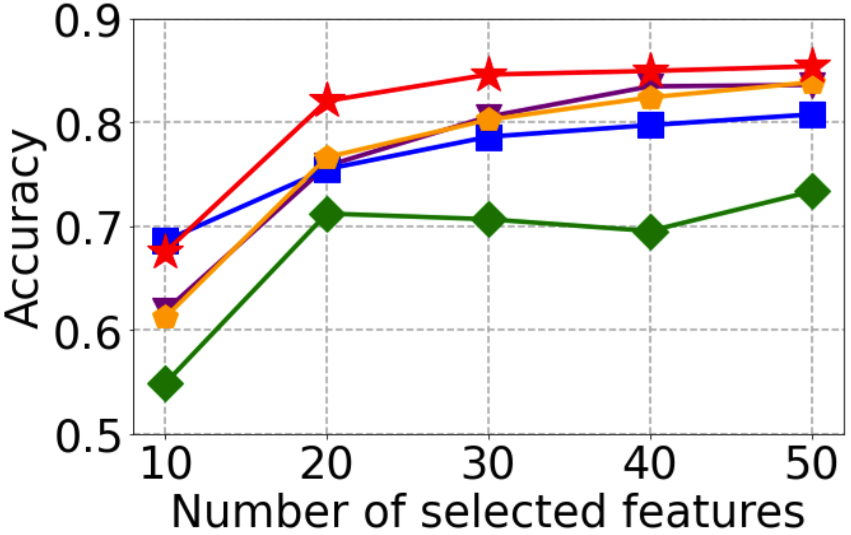}}
\end{minipage}%
}%
}%
\hspace{0.06in}
\subfigure[ALLAML]{
\centering
{
\begin{minipage}[t]{0.3\linewidth}
\centering
\centerline{\includegraphics[width=0.9\textwidth]{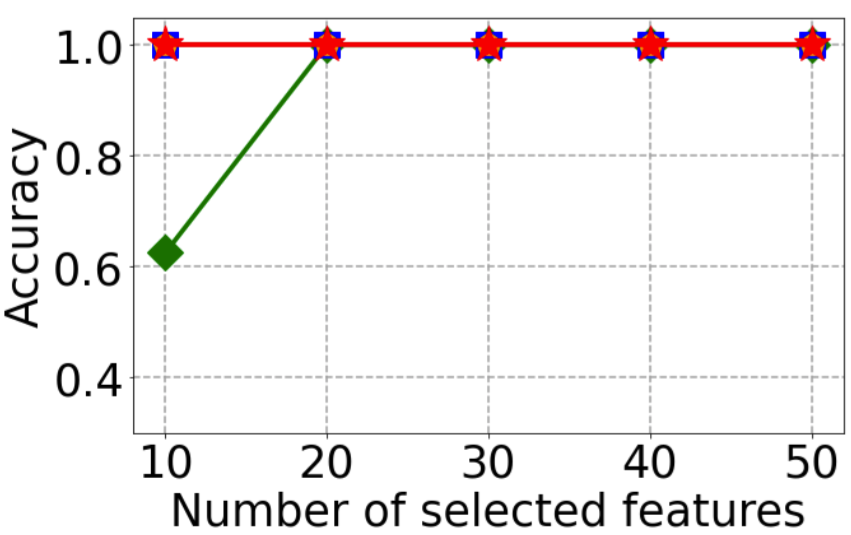}}
\end{minipage}%
}%
}%
\hspace{0.06in}
\subfigure[COIL20]{
\centering
{
\begin{minipage}[t]{0.3\linewidth}
\centering
\centerline{\includegraphics[width=0.9\textwidth]{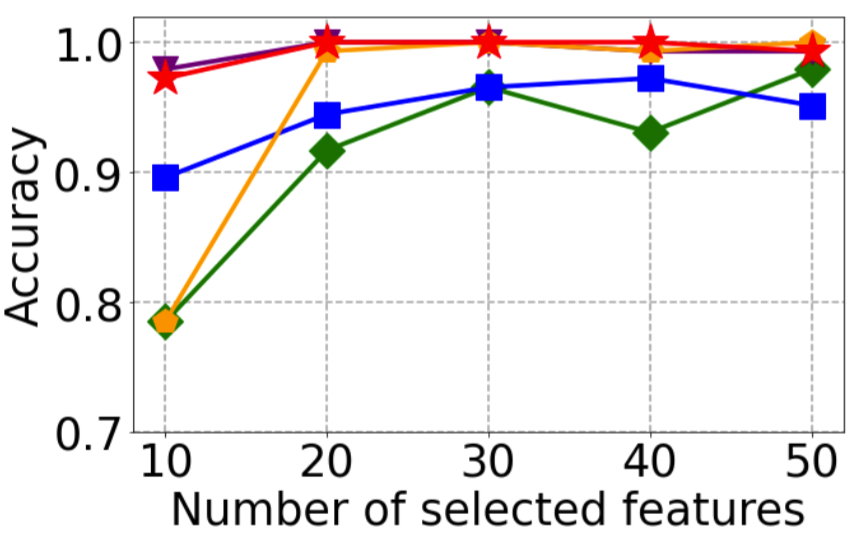}}
\end{minipage}%
}%
}%
\hspace{0.06in}
\subfigure[USPS]{
\centering
{
\begin{minipage}[t]{0.3\linewidth}
\centering
\centerline{\includegraphics[width=0.9\textwidth]{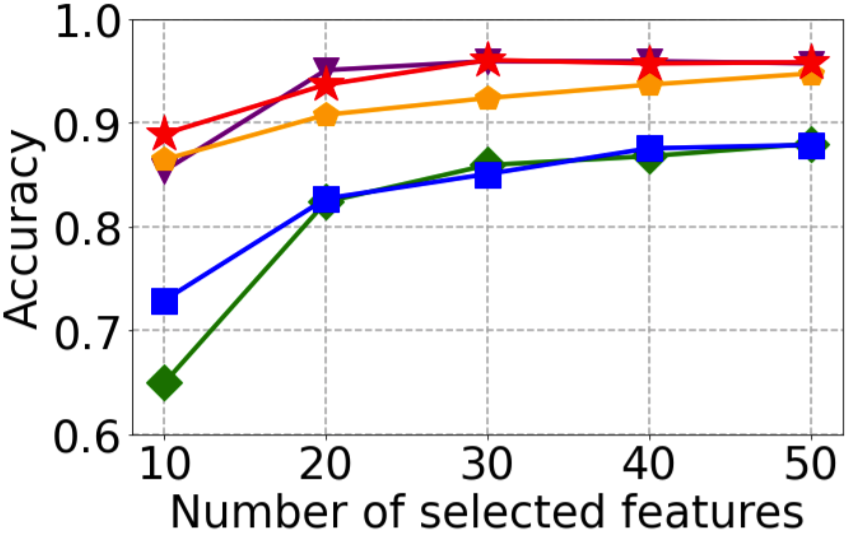}}
\end{minipage}%
}%
}%
\end{center}
\caption{Accuracy versus the number of selected features $k$ for classification.}
\label{classificationRes1}
\end{figure}

To assess the effectiveness of the top-$k$ regularization on classification, we compare~\eqref{DNNClassification} to cutting-edge methods. We vary $k$ from $10$ to $50$ with a step size of $10$ on $6$ datasets and obtain the accuracies of the downstream classifier, i.e., extremely randomized trees. We plot the curves of accuracy versus $k$ in Figure~\ref{classificationRes1}. It is evident that~\eqref{DNNClassification} is consistently better than strong baselines for almost all $k$ values. Further, we average the results for different $k$'s and summarize the results in Table~\ref{classificationRes2}. These results show that~\eqref{DNNClassification} is better and more stable than strong baselines.

\begin{table}[!thp]
\caption{Comparison of feature selection methods in averaged classification accuracy (\%).}
\label{classificationRes2}
\begin{center}
\begin{small}
\setlength{\tabcolsep}{2.8mm}{
\begin{tabular}{lcccccc}
\hline\hline
Dataset 	        			& Yale & MNIST		  	  &MNIST-Fashion      &ALLAML     & COIL20    & USPS \\
\hline
    DFS  		    			&67.1$\pm$13.2&85.0$\pm$11.9  &77.0$\pm$8.1    &{\bf 100.0$\pm$0.0}   &{\bf 99.3$\pm$0.8} &93.6$\pm$4.1  \\
    RF							&40.0$\pm$4.4&84.4$\pm$8.0  &76.6$\pm$4.4    &{\bf 100.0$\pm$0.0}   &94.6$\pm$2.7  &83.2$\pm$5.5 \\	
    Inf-FS						&5.9$\pm$7.4 &16.9$\pm$2.6  &29.0$\pm$3.3    &57.5$\pm$6.1   &40.8$\pm$7.9  &7.5$\pm$1.5 \\
	STG-FS  		    		&65.9$\pm$6.9 &82.4$\pm$11.0  &76.9$\pm$8.2    &{\bf 100.0$\pm$0.0}   &95.4$\pm$8.5  &91.6$\pm$2.9 \\
	FIR-DL						&32.9$\pm$4.7		&73.9$\pm$11.6	&67.9$\pm$6.6	&92.5$\pm$15.0	&91.5$\pm$6.9 &81.6$\pm$8.5  \\
	DNN$^{\mathrm{top}}$ 		&{\bf 71.8$\pm$9.4}&{\bf 89.4$\pm$8.0}  &{\bf 80.9$\pm$6.8}    &{\bf 100.0$\pm$0.0}   &{\bf 99.3$\pm$1.1}  & {\bf 94.0$\pm$2.7}\\
\hline\hline
\end{tabular}}
\end{small}
\end{center}
\end{table}

\section{Discussion}\label{dis}
We will perform ablation study to analyze the role of the top-$k$ regularization, study its stability, and analyze the empirical time complexity of models using it. Further, we will tighten the bound in Theorem~\ref{Approximationerror}, which can be found in the Supplementary Material.

\paragraph{Ablation study.}

\begin{figure}[!thp]
\begin{center}
\centering
{
\begin{minipage}[t]{0.3\linewidth}
\centering
\centerline{\includegraphics[width=1.525\textwidth]{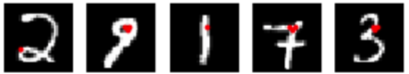}}
\end{minipage}%
}%
\hspace{1.05in}
\centering
{
\begin{minipage}[t]{0.3\linewidth}
\centering
\centerline{\includegraphics[width=1.525\textwidth]{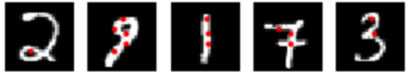}}
\end{minipage}%
}%
\hspace{1in}
\centering
{
\begin{minipage}[t]{0.3\linewidth}
\centering
\centerline{\includegraphics[width=1.5\textwidth]{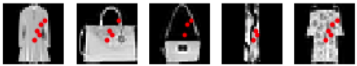}}
\end{minipage}%
}%
\hspace{1.05in}
\centering
{
\begin{minipage}[t]{0.3\linewidth}
\centering
\centerline{\includegraphics[width=1.5\textwidth]{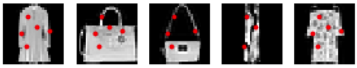}}
\end{minipage}%
}%
\end{center}
\caption{Feature selection with (right panel) or without (left panel) the top-$k$ regularization on MNIST and MNIST-Fashion with $k=5$. Top row: MNIST; Bottom row: MNIST-Fashion.}
\label{ablation1}
\end{figure}

We experiment~\eqref{DNNClassification} and DFS (that is,~\eqref{DNNClassification} without the top-$k$ regularization) on MNIST and MNIST-Fashion, and the results are depicted in Figure~\ref{ablation1} for random samples. It is evident that the features selected by~\eqref{DNNClassification} are more evenly distributed at critical points across each digital image, while the features selected by DFS are more aggregated. This contrastive illustration shows that the top-$k$ regularization helps select informative features without incurring unnecessary redundancy; that is, the top-$k$ regularization promotes the representativeness of selected features.

\paragraph{Representativeness analysis.} 

\begin{figure}[!thp]
\begin{center}
\subfigure[DFS]{
\centering
{
\begin{minipage}[t]{0.16\linewidth}
\centering
\centerline{\includegraphics[width=1.1\textwidth]{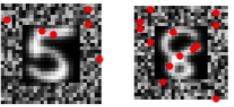}}
\end{minipage}%
}%
}%
\hspace{0.125in}
\subfigure[RF]{
\centering
{
\begin{minipage}[t]{0.16\linewidth}
\centering
\centerline{\includegraphics[width=1.05\textwidth]{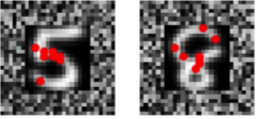}}
\end{minipage}%
}%

}%
\hspace{0.125in}
\subfigure[STG-FS]{
\centering
{
\begin{minipage}[t]{0.16\linewidth}
\centering
\centerline{\includegraphics[width=1.05\textwidth]{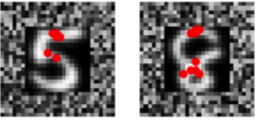}}
\end{minipage}%
}%
}%
\hspace{0.125in}
\subfigure[FIR-DL]{
\centering
{
\begin{minipage}[t]{0.16\linewidth}
\centering
\centerline{\includegraphics[width=1.05\textwidth]{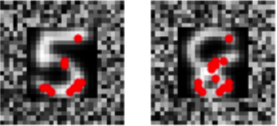}}
\end{minipage}%
}%

}%
\hspace{0.125in}
\subfigure[DNN$^{top}$]{
\centering
{
\begin{minipage}[t]{0.16\linewidth}
\centering
\centerline{\includegraphics[width=1.05\textwidth]{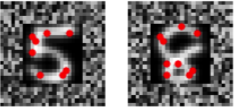}}
\end{minipage}%
}%

}%
\end{center}
\caption{Representativeness analysis of different methods with $k=20$ on USPS.}
\label{ablation2}
\end{figure}

Further, the representativeness of the selected features is analyzed. We resize the original $16\times16$ images into $28\times 28$ images by extending each of the boundaries by $6$ pixels, then we apply different feature selection methods on these images with noise boundaries. Randomly sampled images are illustrated in Figure~\ref{ablation2}. It is found that features selected by DFS are located in the noise edge regions in Figure~\ref{ablation2} (a), while all those selected by~\eqref{DNNClassification} are at salient points evenly distributed across the image parts in Figure~\ref{ablation2} (e). Also, the features selected by other baselines are more aggregated and less representative than~\eqref{DNNClassification}. These results indicate that the top-$k$ regularization helps enhance the representativeness of selected features.

\paragraph{Stability analysis.}

\begin{figure}[!thp]
\centering
{
\begin{minipage}[t]{1\linewidth}
\centering
\centerline{\includegraphics[width=0.35\textwidth]{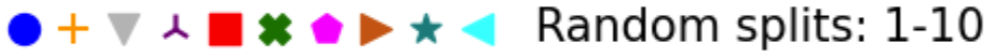}}
\end{minipage}%
}%
\hspace{0in}
\begin{center}
\centering
{
\begin{minipage}[t]{0.2\linewidth}
\centering
\centerline{\includegraphics[width=2.36\textwidth]{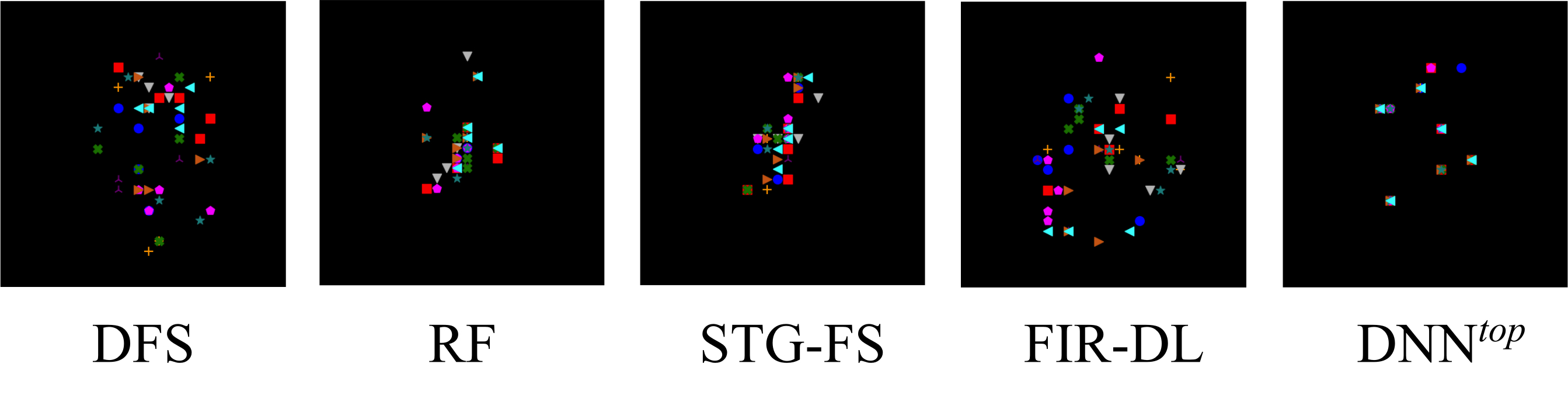}}
\end{minipage}%
}%
\hspace{1.6in}
\centering
{
\begin{minipage}[t]{0.2\linewidth}
\centering
\centerline{\quad\includegraphics[width=2.36\textwidth]{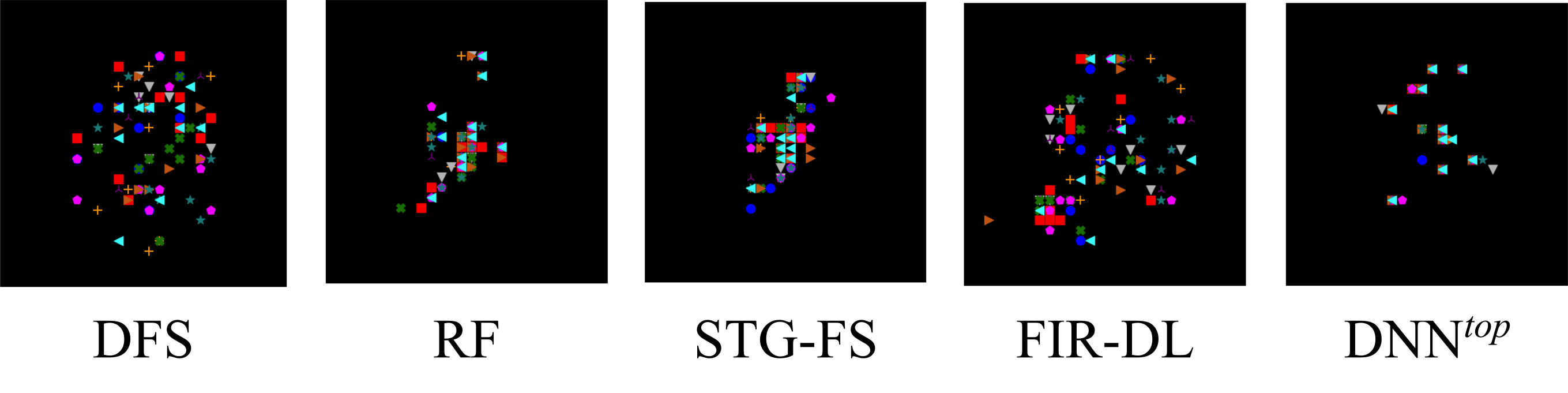}}
\end{minipage}%
}%
\end{center}
\caption{Stability analysis of different methods with 10 random splits for $k=5$ (left panel) and $k=10$ (right panel) on MNIST (best visualized when zooming in).}
\label{Stability1}
\end{figure}

We randomly split the samples of MNIST 10 times, then we use~\eqref{DNNClassification} and strong baselines to perform feature selection. The experimental results are illustrated in Figure~\ref{Stability1}. It is clear that the features selected by~\eqref{DNNClassification} essentially overlap for different splits, whereas those by the baselines do not. Thus, the features from~\eqref{DNNClassification} are more stable compared to those from baselines.

More experimental results on MNIST, MNIST-Fashion, and USPS are provided in the Supplementary Material. Over there, the low-dimensional embeddings of subsets of selected features and all features of MNIST and MNIST-Fashion are also illustrated. It is noted that the embedding of $50$ selected features by~\eqref{DNNClassification} is close to that of all features.

\paragraph{Interpretation from architectural perspective.}

The top-$k$ regularization approach and FIR-DL both require dual-network architectures. However, in FIRL-DL, the two networks are independent, which requires to learn more parameters than training a single network; meanwhile, it is unclear if FIRL-DL is extensible to any other DNN architecture of interest. In stark contrast, the proposed top-$k$ regularization induces a sub-NN that shares weights with the lead NN; also, it is clearly extensible and easily pluggable into a learning model for feature selection. It is worth pointing out that the one-to-one layer to induce the sub-NN is inessential. In the Supplementary Material, we show a flexible extension of the top-$k$ regularization and give a further example of CNN on MNIST; also, we provide the schemas of the top-$k$ regularization for different architectures there.

\paragraph{Computational complexity.}

Experimentally, the computational time of models with the top-$k$ regularization is about twice of the corresponding models without it. This increase of running time is due to the inclusion of the sub-NN; nonetheless, as analyzed in Section~\ref{meth}, the overall computational complexity of models with the top-$k$ regularization is of the same order as the models without it. 


\section{Conclusions}{\label{con}}
In this paper, we propose an effective and concise regularization for supervised feature selection, which can be readily plugged as a sub-model into a given learning model and trained cooperatively. It facilitates a sensible reconciliation of the representativeness and correlations of features, boosting downstream supervised models with these selected features. Theoretically, we analyze the approximation error for using the top-$k$ regularization to approximate high-dimensional sparse functions. Empirically, extensive experiments on real-world datasets demonstrate that the proposed approach has performance on par with or better than strong baseline methods in downstream learning tasks.

\bibliographystyle{abbrvnat}
\bibliography{ref}

\end{document}